\documentclass[letterpapper, 10pt, conference]{ieeeconf} 
\IEEEoverridecommandlockouts
\usepackage{amsmath,amssymb,amsfonts}
\usepackage[utf8]{inputenc}
\usepackage[english]{babel}
\usepackage{algorithmic}
\usepackage{graphicx}
\usepackage{textcomp}
\usepackage{xcolor, colortbl}
\usepackage{csquotes}
\usepackage{svg}
\usepackage{bm}
\usepackage{todonotes}
\usepackage{dblfloatfix}
\usepackage{hyperref}
\hypersetup{
    colorlinks=true,
    linkcolor=black,
    filecolor=magenta,      
    urlcolor=cyan,
    citecolor=black,
    pdftitle={Hands-free teleoperation of a nearby manipulator through a virtual body-to-robot link},
    }

\newcommand{\bx}{{\mathbf{x}}}

\newcommand{\bR}{{\mathbf{R}}}

\newcommand{\bT}{{\mathbf{T}}}

\newcommand{\bdelta}{{\bm{\delta}}}

\newcommand{\fra}[1]{{\mathcal{F}_{#1}}}

\newcommand{\eref}[1]{Eq.~(\ref{eq:#1})}

\newcommand{\nj}[1]{\textcolor{black}{#1}}

\definecolor{blueTransparent}{rgb}{0.8,0.8,1.0}
\definecolor{redTransparent}{rgb}{1.0,0.8,0.8}
\definecolor{purpleTransparent}{rgb}{1.0,0.5,1.0}

\begin{document}
\title{Hands-free teleoperation of a nearby manipulator through a virtual body-to-robot link}

\author{Alexis Poignant$^{1}$, Nathanaël Jarrassé$^{1,2}$, Guillaume Morel$^{1}$
\thanks{$^{1}$Sorbonne Université, CNRS, INSERM, Institute for Intelligent Systems and Robotics (ISIR), Paris, France.}%
\thanks{$^{2}$Email: jarrasse@isir.upmc.fr}%
}

\maketitle

\begin{abstract}
This paper introduces an innovative control approach for teleoperating a robot in close proximity to a human operator, which could be useful to control robots embedded on wheelchairs. The method entails establishing a virtual connection between a specific body part and the robot's end-effector, visually displayed through an Augmented Reality (AR) headset. This linkage enables the transformation of body rotations into amplified effector translations, extending the robot's workspace beyond the capabilities of direct one-to-one mapping. Moreover, the linkage can be reconfigured using a joystick, resulting in a hybrid position/velocity control mode using the body/joystick motions respectively.

After providing a comprehensive overview of the control methodology, we present the results of an experimental campaign designed to elucidate the advantages and drawbacks of our approach compared to the conventional joystick-based teleoperation method. The body-link control demonstrates slightly faster task completion and is naturally preferred over joystick velocity control, albeit being more physically demanding for tasks with a large range. The hybrid mode, where participants could simultaneously utilize both modes, emerges as a compromise, combining the intuitiveness of the body mode with the extensive task range of the velocity mode. Finally, we provide preliminary observations on potential assistive applications using head motions, especially for operators with limited range of motion in their bodies.
\end{abstract}


\section{Introduction}

Body-machine interfaces are devices linked to the body in order to extend or replace body functions \cite{casadio_body-machine_2012}. They include assistive manipulators \cite{jain_assistive_2015}, and generic physical input devices such as mouses, keyboard or computer screens. In the domain of assistance to the disabled, {\it e.g} individuals with spinal cord injury exhibiting residual mobility of upper-body parts \cite{casadio_body-machine_2012}, some devices are controlled through upper-body motions, like head \cite{craig_wireless_2005} or shoulders \cite{casadio_body_2011}. 

\begin{figure}[t!]
\centering
  \includegraphics[width=0.8\linewidth]{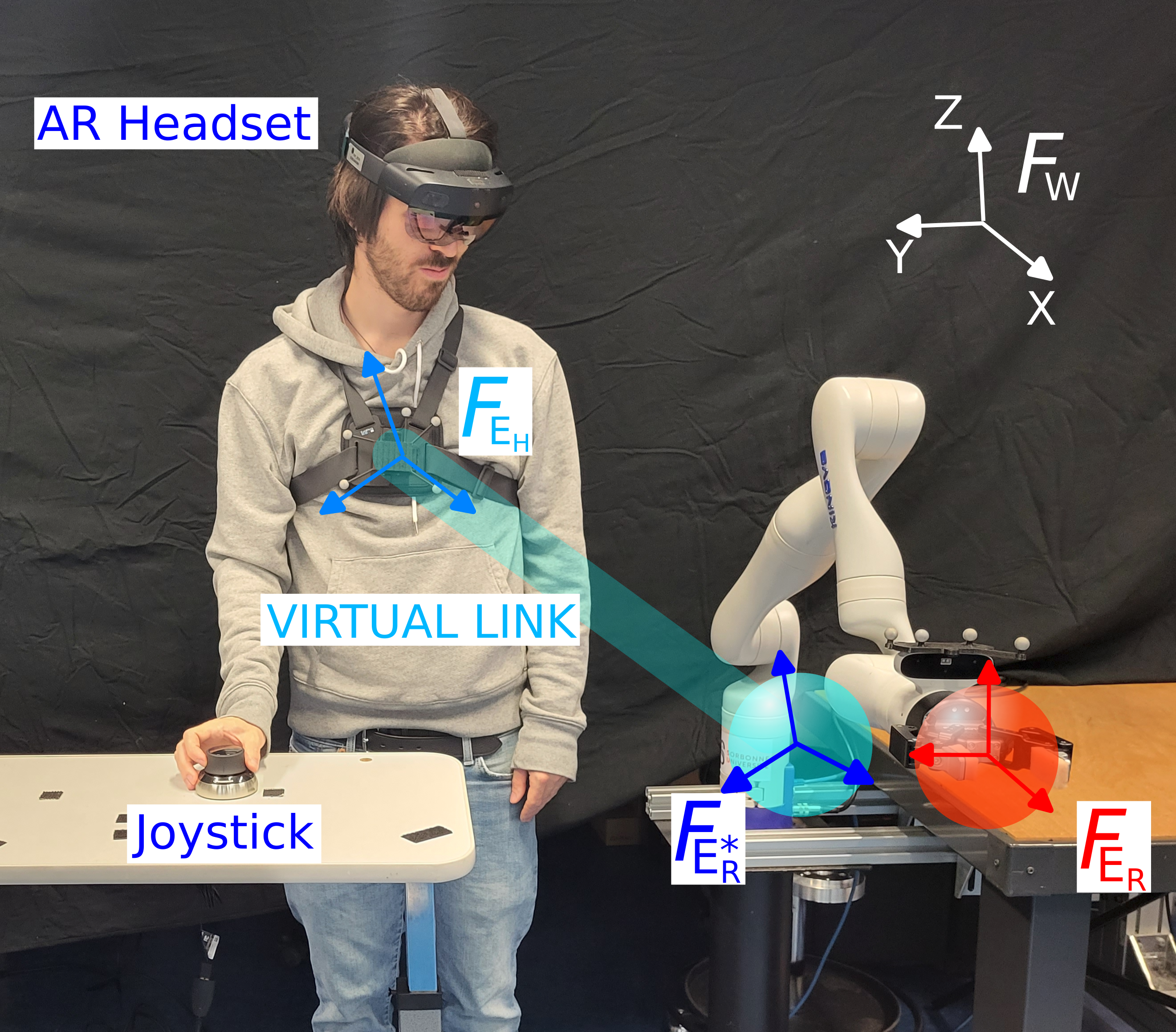}
\vspace{-0mm}
\caption{Representation of the set-up used during the experiments: the virtual end-effector $\fra{{E_R^\star}}$ (light blue sphere) is linked (light blue segment) to the user's thorax $\fra{{E_H}}$, whose movements are captured using Optitrack markers. This link is displayed using a virtual reality headset and can be reconfigured using a 3D joystick manipulated by the user's dominant hand. A robot, set nearby, with end-effector $\fra{{E_R}}$ (red sphere) is then servoed to follow the desired end-effector position. The link and spheres on the image are representations of what is seen through the AR headset. A frame $\fra{{\bullet}}$ is referenced by its three axes $X$, $Y$ and $Z$.}
\label{fig:setup}
\end{figure}

The most common control mode used for these devices is position-to-velocity, where a displacement of a user's body part from a neutral point generates a velocity of the device. Position-to-velocity controllers are particularly efficient for wide-range tasks and delayed interfaces \cite{won_kim_comparison_1987} which make them adapted for wheelchair control for example. However, such velocity control usually require highly customized interfaces \cite{rizzoglio_non-linear_2023} and training.

For manipulation tasks, as long the spatial range is limited, with no significant delay \cite{won_kim_comparison_1987} nor large input noise \cite{couraud_etude_2018}, position-to-position control, which directly maps operator's motions into robotic motions, is therefore often assessed as more intuitive. This approach enables better performances in terms of time \cite{lincoln_visual_1953}, accuracy \cite{jagacinski_fitts_1980} and user-assessed preferences \cite{zhai_human_1995} while requiring no operator's specific tuning as it solely relies on geometric relations. Examples include surgical telemanipulators \cite{freschi_technical_2013} or nuclear or chemical closed spaces \cite{pulgarin_assessing_nodate}, where the environment and the spatial configuration are well controlled. In this master-slave scenario, the mapping between the user's and the robot's displacements is made efficient through a careful alignment between user hands and robot end-effector frames. 

This direct position mapping therefore eases the robot control, but only applies across a small task space and while the operator stays still at a fixed control cabin. This direct mapping is therefore not transposable to assistive manipulators without using amplifying mechanisms like clutching or scaling \cite{dominjon_comparison_nodate} which degrade intuitiveness or precision.

When the robot and the operator are close from each other, another possible position-to-position control approach consists in creating a link between a human body's part and the end-effector. 
Examples include head-pointers, {\it e.g.} mechanical \cite{noauthor_headpointer_2017}\cite{jewell_custom-made_1989} or optical (laser pointers for computer screens, \cite{goodman_computer_2008}). Those are used to convey desired 3D or 2D positions through head movements and could be used to \nj{input} desired positions of a robotic effector. This pointer-mapping, or rotation-to-position mapping, allows to transform body rotations into large effector translations using the leverage of the link. Unlike direct mapping, it scales the operator's motions, without requiring any tuning like traditional scaling or velocity control interfaces as shown in \cite{jewell_custom-made_1989}. Finally, it is also "hands-free" and can, unlike joystick, utilize dexterity of the head or the thorax while the hand might present little to no mobility, especially in rehabilitation scenarios.

For such a mechanical pointer connected to a human body part, the desired robot end-effector location given by the pointer, with respect to the world frame, $\bT_{W \to E_R^\star}$, can be computed as:
\begin{equation}
    \bT_{W \to E_R^\star}(t) = \bT_{W \to E_H}(t) ~~  \bT_{E_H \to E_R^\star} ~,
    \label{eq:equation_pointer}
\end{equation}
where $\bT_{A\to B}$ is the homogeneous  transform from frame $\fra{A}$ to frame $\fra{B}$, $\fra{{E_H}}$ a frame linked to the human body effector, where the pointer base is attached, $\fra{{E_R^\star}}$ is the desired frame location for the robotic end-effector and $\fra{W}$ the world fixed frame. In this equation, $\bT_{W \to E_H}(t)$ shall be measured to record the human motions while $\bT_{E_H \to E_R^\star}$ is constant and reflects the shape of the pointer. Finally, the real robot with effector $\fra{{E_{R}}}$ and base fixed with respect to $\fra{W}$, is servoed to follow $\fra{{E_{R}^\star}}$, with unavoidable delays in the position tracking process.

Using rigid physical pointers (and thus constant $\bT_{W \to E_R^\star}$) limits the attainable task space for the user. In this paper, we propose to implement a virtual pointer whose geometry can be changed using an external control, namely a joystick. With this system, referring to \eref{equation_pointer}, the user can control the desired robot position $E_R^\star$ using whether its own body motions $\bT_{W \to E_H}(t)$ or by reshaping on-line the pointer, making  $\bT_{E_H \to E_R^\star}(t)$ time varying. In order for the user to benefit from a visual feedback of the virtual pointer end-effector's location, $\fra{{E_R^\star}}$, we add in the setup an augmented reality headset that displays the current virtual link shape. The visualization allows to easily deal with the inherent delay of the cobot, which is slower than the operator, especially due to the lever effect.

In the following sections, we first detail the control algorithms and then report experimental results aimed at determining how humans subjects try to achieve positioning tasks with such a system, when they are asked to use one of the two control inputs, or both simultaneously.

\section{Different Modes and Methodology}

\subsection{Experimental set-up}

The set-up is presented in Fig.~\ref{fig:setup}. 
The virtual device base frame $\fra{E_H}$ is materialized by an optical marker attached to the user's thorax. The marker is localized in real time by an Optitrack motion capture system w.r.t. its fixed frame $\fra{O}$. In this paper, the thorax was chosen as a body part that presents limited translation motions, making it therefore unsuitable for direct position-to-position mapping, while having little influence on the perception of the task (unlike the head) or the hand motions, allowing an industrial scenario where the thorax controls a proximal robot, while the hands are performing another task.

A\nj{n illustrative} video detailing the system shows other choices for the location of the marker $\fra{E_H}$  (head, elbow), is available online\footnote{\url{https://www.youtube.com/watch?v=EgwzT784Fws}}. The experiment also involves a Hololens 2 Augmented Reality (AR) headset that is self-localized in real-time w.r.t. a fixed $\fra{W_{AR}}$, and a 7 DOFs Kinova GEN3 robot with fixed base frame $\fra{B_R}$. It is noteworthy that in a laboratory setting, the motion capture system offers precise and high-frequency tracking. However, we highlight that tracking of the human frame can also be accomplished using the native headset SLAM (Simultaneous Localization and Mapping) and inertial measurement units, as demonstrated towards the end of the video.

Prior to the experiment, a procedure allows for registering $\fra{O}$, $\fra{W_{AR}}$ and $\fra{B_R}$, all with respect to a fixed world frame $\fra{W}$ (chosen arbitrarily). This allows to express in the same frame $\fra{W}$ the optical measurements, the location of the pointer tip to be displayed in the AR headset, and the robot position or velocity commands.

The AR headset is used to display the pointer end-effector location to the participant. This end-effector is a sphere centered at the origin of frame $\fra{E_R^\star}$, whose orientation is kept constant during an experiment. Namely, the desired pose for the robot writes:
\begin{equation}
\bT_{W \to E_R^\star}(t) = 
    \begin{pmatrix}
    \bR_{W \to E_R}(t_0) & \bx_{W \to E_R^\star}(t) \\
    0~~~~0~~~~0 & 1
    \end{pmatrix}~,
    \label{eq:desired_position_to_desired_pose}
\end{equation}
where $\bR_{A \to B}$ and $\bx_{A \to B}$ are the rotation matrix and the origin translation from $\fra{A}$ to $\fra{B}$, respectively, while $t_0$ is the time at the beginning of the experiment.

The robot end-effector location $\bT_{W \to E_R}$ is servoed to $\bT_{W \to E_R^\star}(t)$ using a simple resolved rate controller imposing an end-effector velocity proportional to the error. The controller guarantees, due to its integral effect, a null permanent error. The convergence rate is tuned thanks to a simple proportional gain $k$ determining the closed-loop position control bandwidth. Its value is set to $k=0.5~s^{-1}$, heuristically determined to obtain the fastest behavior before exciting the internal joint position controller oscillations.

In a closely proximate scenario, such as when the robot is affixed to a wheelchair, the controller necessitates implementation with kinematic constraints. These constraints can be realized through the utilization of bounding boxes that prohibit the robot from entering specific regions, like the participants themselves. One effective method for achieving this is by employing a Quadratic Programming (QP) controller, as demonstrated in \cite{khoramshahi_practical_2023}. During the subsequent experimental campaign, where participants were situated at a given distance from the robot, the inverse kinematics were solved using a conventional pseudo-inverse Jacobian matrix.

\subsection{Teleoperation control modes}

Three teleoperation modes were implemented to allow participants controlling the desired position $\bx_{W \to E_R^\star}(t)$.

\subsubsection{Joystick Mode}
With the joystick mode, the user controls the robot desired velocity thanks a 3D SpaceMouse joystick. This corresponds to state of the art and will be used as a reference. The robot end-effector desired pose is computed thanks to \eref{desired_position_to_desired_pose}, with: 
\begin{equation}
    \bx_{W \to E_R^\star}(t) = \bx_{W \to E_{R}}(t_0) + \int_{t_0}^{t} \dot{\bx}^J_{W}(\tau)d\tau,
\label{eq:joystick_control}
\end{equation}
with $\dot{\bx}^J_W$ the velocity input provided by the joystick at all time relative to the reference world frame $W$.

The velocity is computed from \nj{the} joystick output with a gain that tunes the sensitivity of the system. Prior to the recorded experiments, each participant performs a few positioning experiments with several gains, and chooses the preferred one. The joystick is a 3D SpaceMouse compact \cite{noauthor_spacemouse_nodate}, which is a 6 degrees of freedom joystick, though only its translations where used in this experiment. In this mode, the virtual effector $E_{R}^\star$ is displayed in the AR headset as a sphere. Controlling the virtual end-effector $E_{R}^\star$ and not the real effector allows to have velocity inputs that are more important than without, and deletes the potential delay of the velocity control loop and maximum acceleration of the robot at the start of the motions.

\subsubsection{Body Mode}
In the body mode, the user mobilizes his/her thorax movements to move the robot end-effector thanks to a rigid virtual link between $\fra{E_H}$ and $\fra{E_{R}^\star}$. Namely:
\begin{equation}
    \bx_{W \to E_R^\star}(t) = \bR_{W \to E_H}(t) ~ \bx_{E_H \to E_R}(t_0) + \bx_{W \to E_H}(t)~.
    \label{eq:body_control}
\end{equation}

At the beginning of each experiment, the robot end-effector is at a given initial location $\bT_{W \to E_R}(t_0)$ while the participant is asked to stand at a given location (without high precision). Its precise location $\bT_{W \to E_H}(t_0)$ is recorded and used to shape the pointer $\bx_{E_{H} \to E_R^\star}(t_0)$.
The shape of the pointer might vary depending on the physiognomy of the participants, but ensures that $E_R$ is in the center of the robot task space, \nj{to avoid getting} close to singularities. The length of the link was around 1~meter for all the participants. In order to move the effector, participants could either translate their thorax (they were authorized to put one of their foot forward or backward for stability, but not both), or rotate it.

\subsubsection{Dual Mode}
In the dual mode, the user can use simultaneously the joystick and his/her thorax movements to move the desired end-effector:
\begin{equation}
    \bx_{W \to E_R^\star}(t) = \bR_{W \to E_H}(t) ~  \bx_{E_{H} \to E_{R}^\star}(t) + \bx_{W \to E_H}(t),
\label{eq:hybrid_control_1}
\end{equation}
where $\bx_{E_{H} \to E_{R}^\star}(t)$ depends on the joystick's inputs:

\begin{eqnarray}
        \bx_{E_H \to E_R^\star}(t)& =&  \bx_{E_{H} \to E_{R}}(t_0)  \nonumber \\
        &&+ \int_{t_0}^{t} \bR_{E_H \to W}(\tau) ~~ \dot{\bx}^J_{W}(\tau)d\tau~~. \label{eq:hybrid_control_2}
\end{eqnarray}

Notice that, if the participant stays still, then  $\forall t, \bT_{W\to E_H}(t)=\bT_{W\to E_H}(t_0)$ the hybrid controller given in Eqs.~(\ref{eq:hybrid_control_1}-\ref{eq:hybrid_control_2}) is equivalent to the joystick controller given in \eref{joystick_control}. Similarly, if the participant does not use the joystick and teleoperates the robot from his/her body movements only, then $\forall t, \dot{\bx}^J_{W}(t)=0$, and the hybrid controller is equal to the body control mode described in \eref{body_control}.

\subsection{Protocol}
Before starting the  experimental session, subjects could train as long as they wished using the joystick and familiarizing with the virtual+robotic environment, without being asked to precisely reach any target. This familiarization session lasted typically five minutes. The sensitivity of the joystick could be adjusted by the subjects according to their preference. All of them chose a sensitivity gain leading to a maximum velocity of the virtual effector $E_R^\star$ faster than the time response of the position controller used to servo $E_R$, thus the utilization of the AR even in joystick mode. There was no preliminary training with the Body mode nor Dual mode.

The whole experiment corresponded to 7 trials that the participants had to perform using the three different control modes, in the following order: 
\begin{itemize}
    \item 1st, 4th, 7th trials using Joystick mode (Joystick~1, 2 and 3 respectively),
    \item 2nd, 5th trials using Body mode (Body~1 and 2),
    \item 3rd, 6th trials using Dual mode (Dual~1 and 2).
\end{itemize}
Subjects rested as much as they wanted  between trials. The joystick was performed 3 times as we suppose that on the first trial people were \nj{still} not fully familiarized with AR headset and visualization. Therefore, a third joystick trial was added as a final reference.

\begin{figure}[bht]
\centering
  \includegraphics[width=0.35\linewidth]{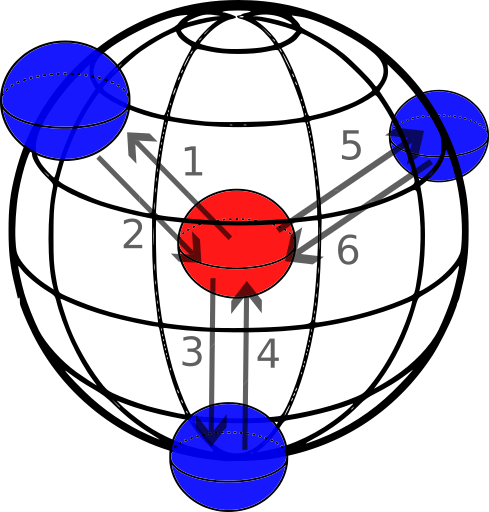}
\vspace{-2mm}
\caption{A task example with 3 blue targets and 6 trajectories (numbered arrows). The task starts from the central point (red sphere), goes to one blue spherical target centered on the black sphere surface, and then back to center. Each target is completed  when the robot end-effector $E_R$ has stayed for 1s less than 2cm away from the target center. Only 1 target appears at a time. Real trials consisted of 15 blue targets to reach.}
\label{fig:task}
\end{figure}

Each of the 7 trials consisted of reaching 30 targets, 15 of them being the same central point, 15 others spread on a 15 cm radius sphere following the Fibonacci distribution, as outlined in \cite{gonzalez_measurement_2010}. This distribution facilitates the exploration of directions that are nearly uniformly distributed on the sphere. The target alternated between the central point and the surface of the sphere, creating 15 back and forth trajectories (see Fig.~\ref{fig:task}). 

Each target reaching was acknowledged when the robot stood at least 1 second inside a 2cm radius tolerance sphere. This tolerance was chosen such as the real robot effector had to immobilize close to the target, but such as it was not too difficult for the subjects to maintain (mostly due to the depth being hard to distinguish for some participants in the AR visualization, as well as their slight sway or tremor). 

Total completion of all the trials took from 30 minutes to 1 hour depending on the participants' performances and rest sessions duration.

\subsection{Metrics}
From a first set of preliminary experiments with five subjects (notice that the results of \nj{those} first five participants were not included in the experimental results analysis), we analyzed the signals (participant and robot movements, joystick output) and selected three indexes:

\subsubsection{Completion time per target} Since the task was the same for all the control modes, we did not study the precision but rather the duration to complete each target, from when it appeared to its validation.

\subsubsection{Total Body Displacement per target}
In order to evaluate how much a participant uses his/her body during a task, we measure the displacement between the beginning of the task (at $t_0$, when the target appears in the AR headset) and its validation at $t_f$ (task completed), $\bT_{E_H(t_0) \to E_H(t_f)}$.

We then evaluate the following 6 components ($X$,$Y$,$Z$ in translation, $\alpha$,$\beta$,$\gamma$ in rotation) of this displacement:

\begin{equation}
\left ( \begin{array}{c}
     \delta_{H_X}  \\
     \delta_{H_Y}  \\
     \delta_{H_Z} \\
     \delta_{H_\alpha}  \\
     \delta_{H_\beta}  \\
     \delta_{H_\gamma}
\end{array} 
\right )
= \left ( \begin{array}{c}
~~\\\left | \bx_{E_H(t_0) \to E_H(t_f)} \right |\\
~~\\~~\\
\left | \text{angleaxis}\left (\bR_{E_H(t_0) \to E_H(t_f)} \right ) \right |\\ 
~~ 
\end{array} 
\right )~,
\label{eq:def_delta_H}
\end{equation}
where angleaxis$(\bR)$ is a vector oriented along the rotation axis of $\bR$ with a norm corresponding to its geodesic distance. X corresponds to the trunk moving forward/backward, Y sideways and Z up and down.

\subsubsection{Control distribution in the Dual Mode} 
During trials 3 and 6 (dual mode) participants can perform the task using only the joystick controls, only body movement controls or a mix of both. Joystick controls modify the shape of the virtual link between the body marker and the desired robot location, namely $\bx_{E_H \to E_R^\star}(t)$. The contribution to the robot movement produced by the joystick input, from $t_0$, at a given time $t$, writes:

\begin{equation}
    \boldsymbol{\delta}_{J}(t) = \bx_{E_H \to E_R^\star}(t) - \bx_{E_H \to E_R^\star}(t_0)~.
\end{equation}

The total robot displacement (produced by both inputs), from $t_0$ (was resetted each time a goal was attained) to $t$ expressed in the initial body frame $\fra{E_H}(t_0)$, writes:

\begin{equation}
    \boldsymbol{\delta}(t) = \bR_{E_H \to W} (t_0) \left (  \bx_{W \to E_R^\star}(t) - \bx_{W \to E_R^\star}(t_0) \right )~.
\end{equation}
We therefore define the contribution to the robot displacement by the body control mode by:

\begin{equation}
    \boldsymbol{\delta}_{B}(t) = \bdelta(t)-  \boldsymbol{\delta}_{J}(t)
\end{equation}

Finally for the analysis, we use the contribution of each control mode to the robot movement normalized over the total displacement. Namely, along the $\bullet$ axis (with $\bullet$ corresponding to $X$, $Y$ or $Z$), we define the following ratios:

\begin{equation}
\left \{    
\begin{array}{rcl}
     b_{\bullet}(t) & = & \displaystyle \frac{{\delta}_{B_\bullet}(t)}{{\delta}_{\bullet}(t_f)}  \\ [8 pt]
     j_{\bullet}(t) & = & \displaystyle \frac{{\delta}_{J_\bullet}(t)}{{\delta}_{\bullet}(t_f)}
\end{array}
\right. ~,
\end{equation}
where $\bdelta(t)=\left [ \delta_X~~\delta_Y~~\delta_Z\right ]^T$, $\bdelta_B(t)=\left [ \delta_{B_X }~~\delta_{B_Y}~~\delta_{B_Z}\right ]^T$ and $\bdelta_J(t)=\left [ \delta_{J_X}~~\delta_{J_Y}~~\delta_{J_Z}\right ]^T$, all in $\fra{E_H}(t_0)$.

In any direction $X$, $Y$ or $Z$, data points where the total displacement of the effector in that direction was inferior to 2 cm (task tolerance) were not considered in order to remove noisy measures from the distributions.

\subsection{Questionnaires}
Additionally, after the whole experimental session, participants were asked to fill questionnaires with six questions. From French to English they can be translated as:
\begin{enumerate}
    \item To what extent has the task required a physical effort?
    \item To what extent has the task required a cognitive effort?
    \item How much did you feel in the control of the robotic arm?
    \item How intuitive was it to control the robotic arm?
    \item To what extent have you felt that the robot was an extension of yourself?
    \item Which global score would you rate this experiment?
\end{enumerate}

Participants answered with a score ranging from -3 (very little) to 3 (very much).

One questionnaire was filled for each control mode. For the reading of the results in the next section it is worth remembering that a high rated answer for questions~1 and~2 corresponds to a negative characteristics of the system (high load), while, for questions~3 to~6 it reflects a positive characteristics (intuitiveness, general satisfaction, etc.).

\subsection{Participants}
The experimental study was carried out in accordance with the recommendations of Sorbonne Universit\'e ethics committee CER-SU, which approved the protocol. 

Fourteen asymptomatic participants, aged 20-45, volunteered for this experimental study. 
They all gave their informed consent, in accordance with the Declaration of Helsinki.

\section{Results}
Data presented in the results are not normal according to the Shapiro-Wilk test. Therefore the Mann-Withney-U Test was used to compute p-values and to analyse statistical difference between sessions and modes. In figures, $\star$ denotes significant difference at $p < 0.05$, and $\star\star$ at $p < 0.005$. When 2 comparisons where made using the same data, a Bonferroni correction of 2 was applied.

\subsection{Performances and preferences between modes}
\begin{figure}[h!]
\centering
  \includegraphics[width=0.60\linewidth]{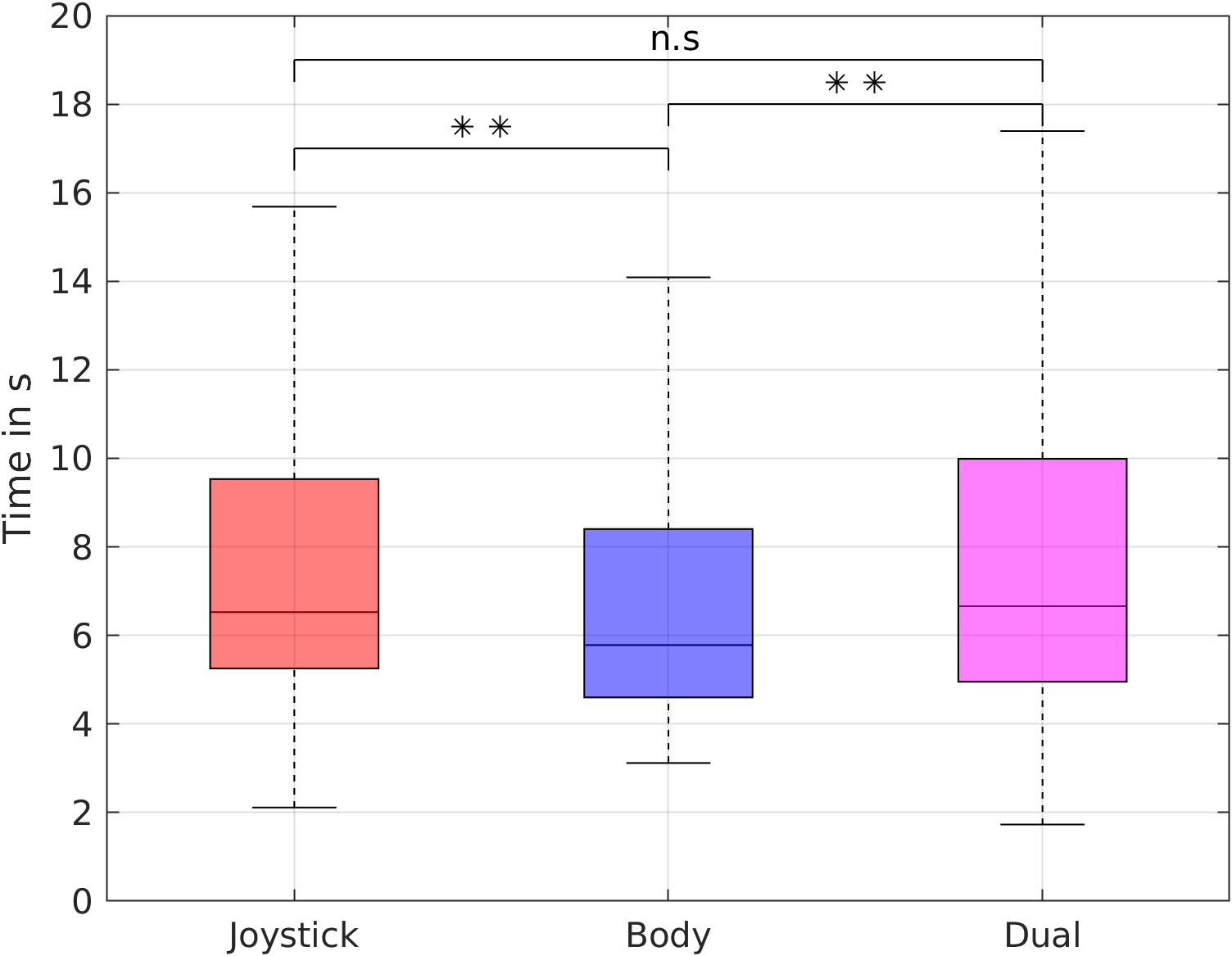}
\vspace{-2mm}
\caption{Distribution of the completion time per target for each participant (medians, 25th and 75th quartiles, and whiskers of width 1.96 the standard.}
\label{fig:time}
\end{figure}

A distribution of the completion time per target for all the participants is provided \nj{in} Fig.~\ref{fig:time}. Body mode was the fastest, approximately 10\% than the other modes ($p < 0.005$), and with a lower variance. 

\begin{figure}[h!]
\centering
  \includegraphics[width=0.60\linewidth]{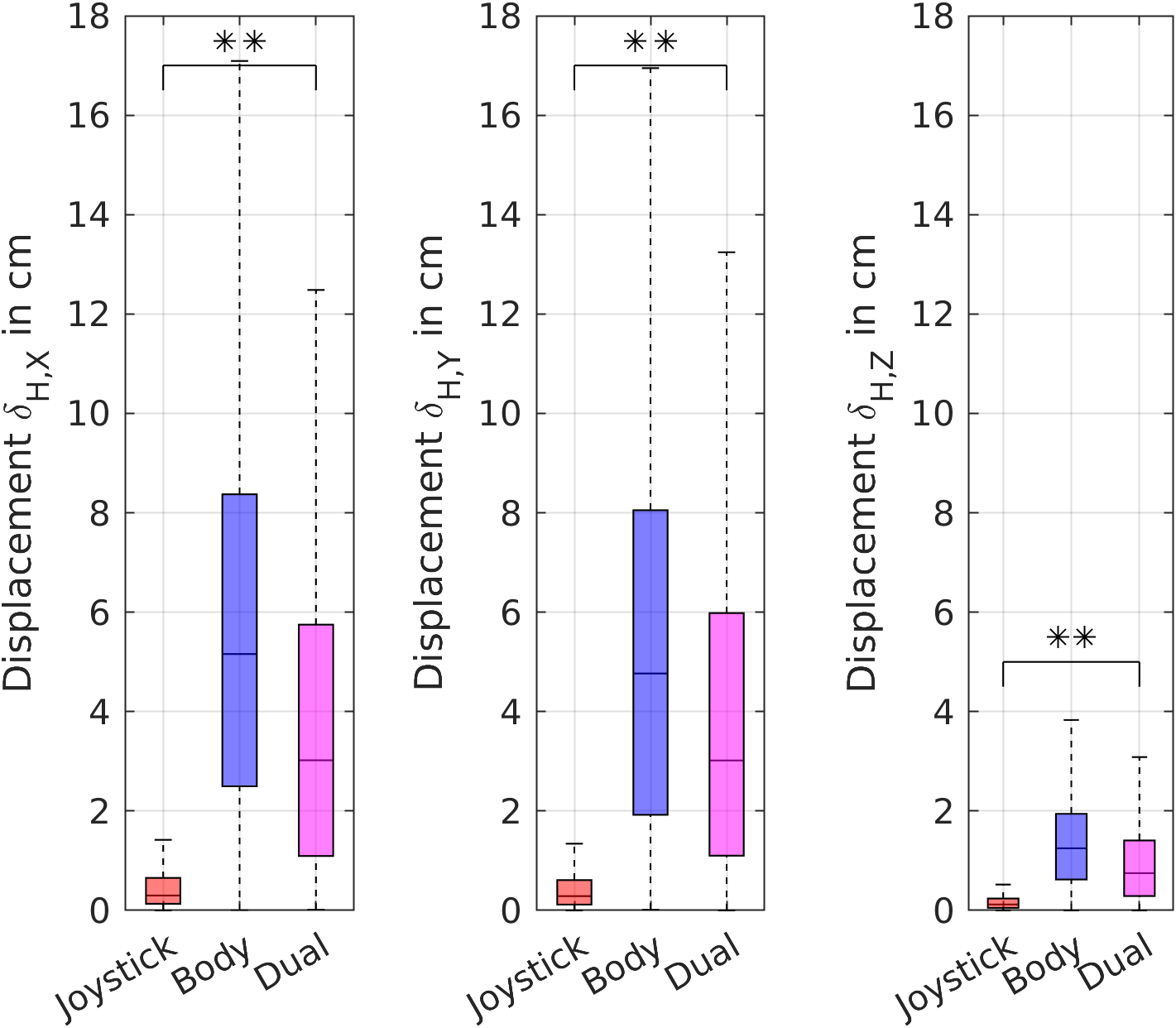}
\vspace{-2mm}
\caption{TBD of the thorax per target in translation in cm}
\label{fig:rangeTrans}
\end{figure}

\begin{figure}[h!]
\centering
  \includegraphics[width=0.60\linewidth]{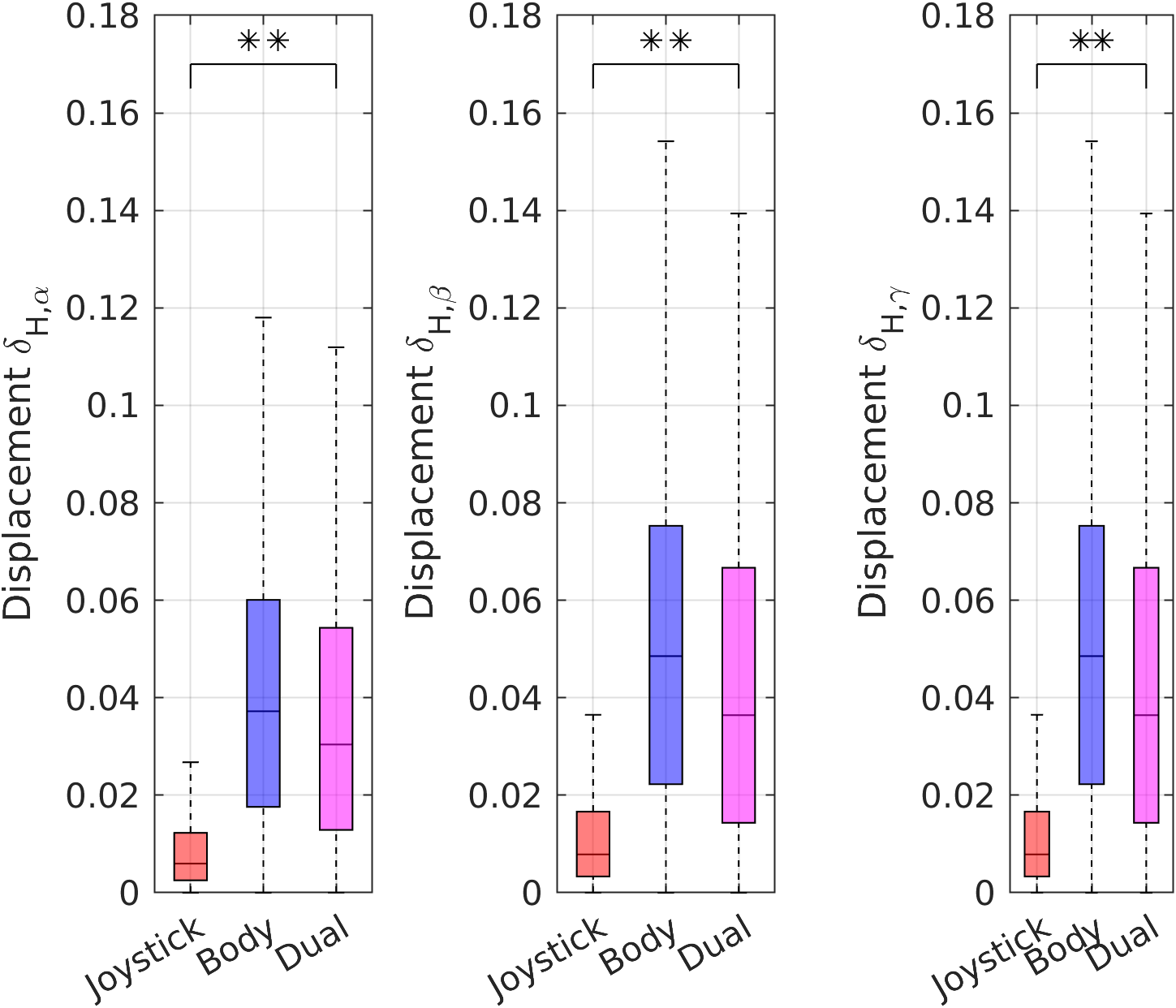}
\vspace{-2mm}
\caption{TBD of the thorax per target in rotation in radians}
\label{fig:rangeRot}
\end{figure}

The total body displacement (TBD) of the thorax per target along the three axes in translation and rotation are provided \nj{in} Fig.~\ref{fig:rangeTrans} and \ref{fig:rangeRot} respectively.
In any mode, $\delta_{H_X}$ and  $\delta_{H_Y}$ are much higher than  $\delta_{H_Z}$. Further, the rotation $\delta_{H,\alpha}$ around the X-axis is less used than $\delta_{H,\beta}$ and $\delta_{H,\gamma}$. Nonetheless, in all directions, both in translation and rotation, the displacements exhibited by the participants' were significantly higher in Body Mode compare to Joystick Mode, around 3 to 6 times higher depending on the axes. Similarly, Dual Mode exhibits values around 3 to 4 times higher than \nj{the ones} in Joystick Mode, but significantly lower than in Body Mode.

\begin{figure}[h!]
\centering
  \includegraphics[width=0.8\linewidth]{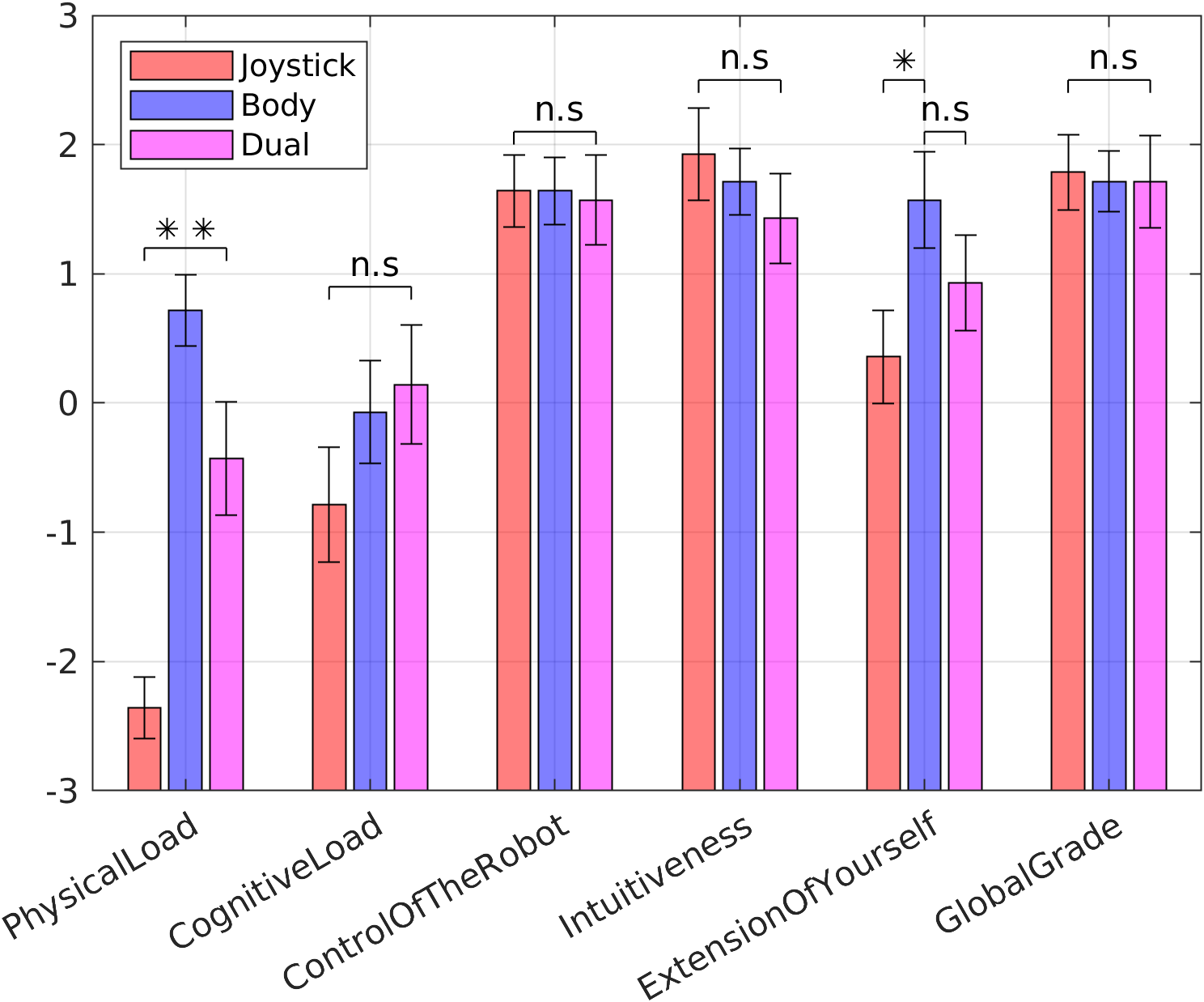}
\vspace{-2mm}
\caption{Questionnaires responses distribution for each mode along the 14 participants.}
\label{fig:questionnaireSphere}
\end{figure}

Finally, the questionnaire results shown Fig.~\ref{fig:questionnaireSphere} exhibit that:
\begin{itemize}
    \item the \nj{average} global grade was equally positive for all control modes (+1.8);
    \item the body control has been rated significantly more physical than the other modes ($p < 0.005$). Dual mode is also significantly less physical than body mode;
    \item there are no significant differences between all the modes regarding questions 2, 3 and 4 ($p > 0.2$ for all the tests);
    \item during the body control mode, the arm is rated as slightly more as an "extension" of oneself ($p < 0.05$).
\end{itemize}

\subsection{Control distribution during dual mode}

\begin{figure}[h!]
\centering
  \includegraphics[width=0.8\linewidth]{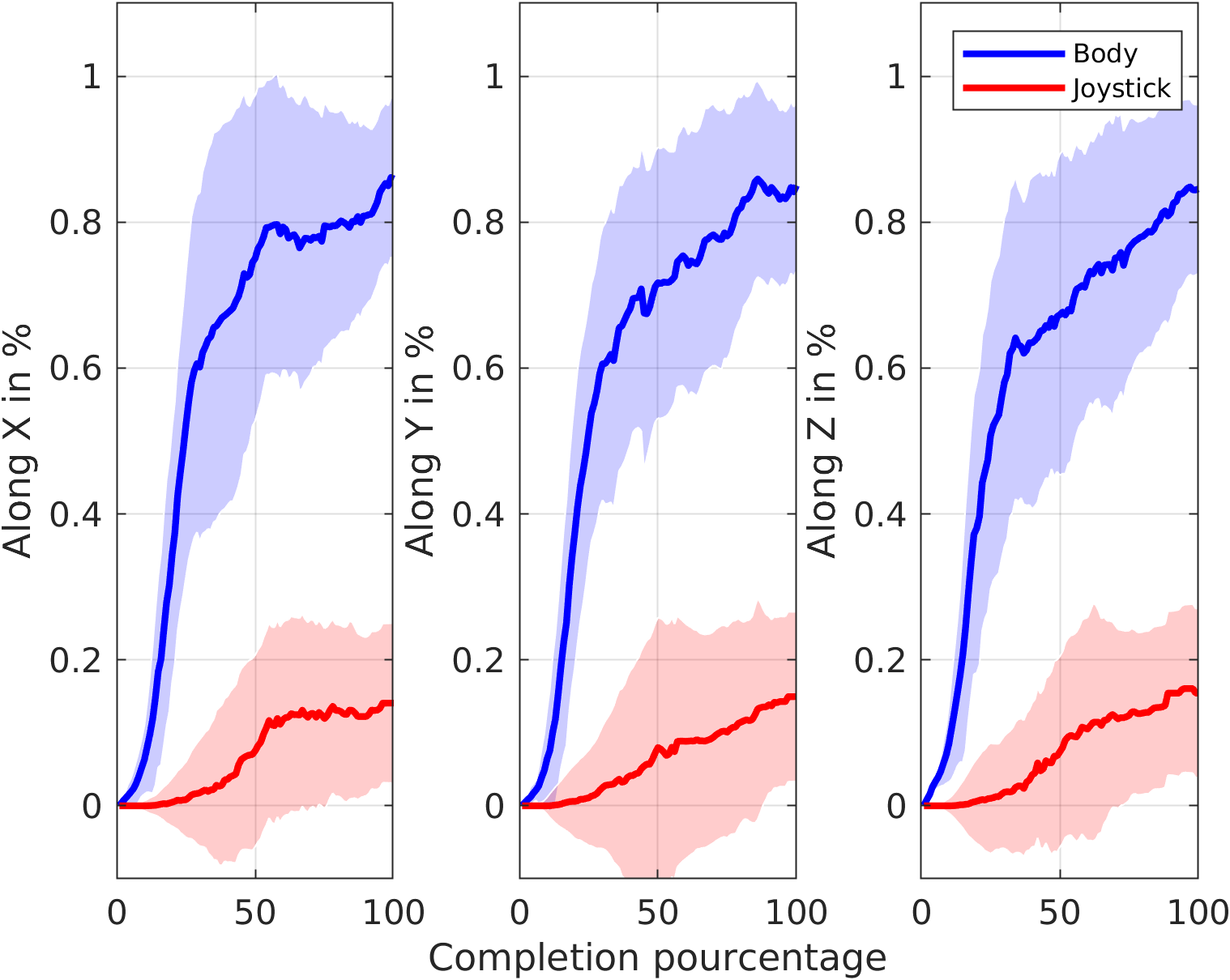}
\vspace{-2mm}
\caption{Median trajectory curve of the control distribution in dual mode, for both the body $b_{\bullet}(t)$ (blue) and joystick $j_{\bullet}(t)$ (red), with standard deviation in a lighter color.}
\label{fig:bodyUsageCurve}
\end{figure}

In Dual mode, the median curve of the body $b_{\bullet}(t)$ contribution and joystick contribution $j_{\bullet}(t)$ relative to the whole robot displacement, $b_{\bullet}(t_f)$ are shown Fig.~\ref{fig:bodyUsageCurve}. Overall, in all directions, body movements contributed to 80\% of the task in median (and therefore the joystick for the remaining 20\%), with the upper 75th quartile being close to 100\%. We can also see that the \nj{end-effector} motions due to the body and joystick are sequential: the body engenders a fast and wide motion at start, then, in a second step, body and joystick simultaneously adjust the desired position.

\subsection{Evolution of the performances across trials}

\begin{figure}[h!]
\centering
  \includegraphics[width=0.65\linewidth]{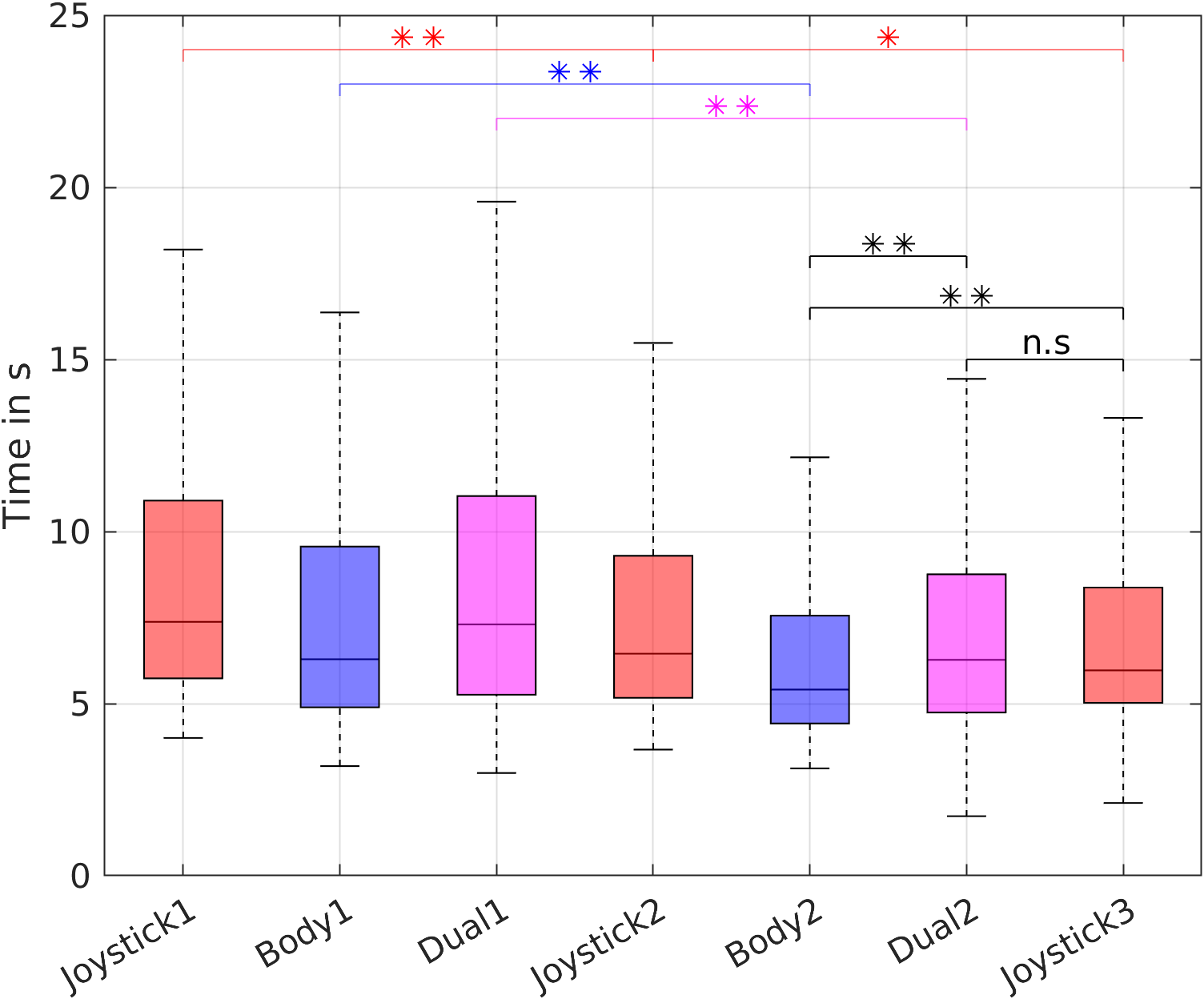}
\vspace{-2mm}
\caption{Distribution of the completion time per target per trial for each participant (14 participants, 30 targets each)}
\label{fig:times}
\end{figure}

A distribution of the time of completion per trial is shown in Fig.~\ref{fig:times}. Between the first and second trial (and third trial for joystick) of each mode, we observed a lower median duration with a statistically significant difference. Dual2 and Joystick3 presented similar median times (6.3 and 6.0~s respectively, with no statistically significant difference) while Body2 was slightly faster with a lower median time (5.4~s) and a smaller variance. This is consistent with the results presented in Fig.~\ref{fig:time}.

\begin{figure}[h!]
\centering
  \includegraphics[width=0.65\linewidth]{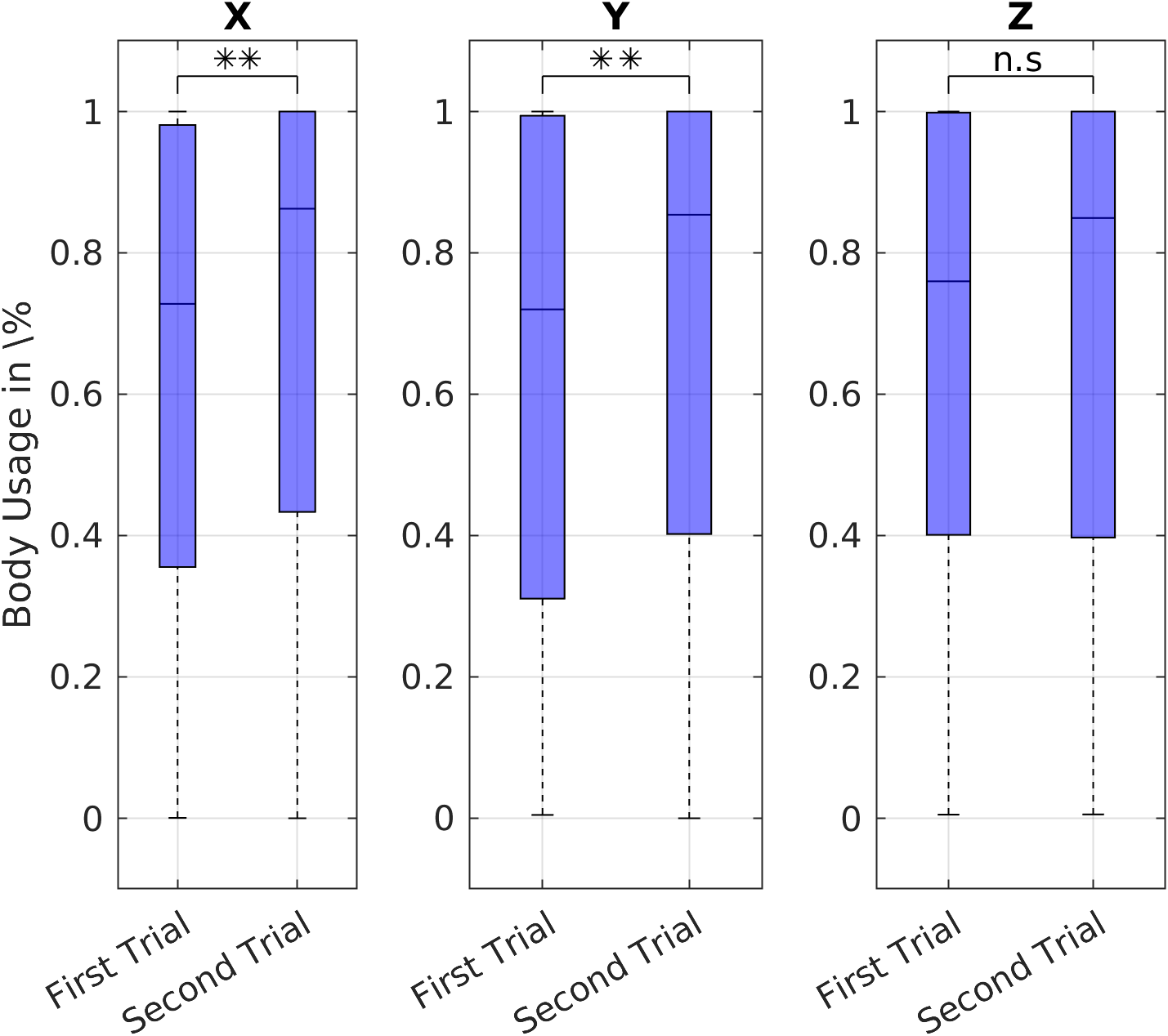}
\vspace{-2mm}
\caption{Body usage in percent regarding the task completion (Medians: X: 0.73 and 0.86 / Y 0.72 and 0.85 / Z 0.76 and 0.85)}
\label{fig:bodyUsage}
\end{figure}

We can also look at $ b_{\bullet}(t_f)$ between both of the dual trials, shown in Fig.~\ref{fig:bodyUsage}. \nj{Along} X and Y, we can see \nj{that} between the first and second trial this percentage has increased significantly, going from 75\% to 85\%. \nj{Along} the Z direction, the first trial is closer to 80\% and does not increase significantly after ; this might be due to the fact that people were already using more body motions \nj{along} Z as they are not used to Z-joystick and more accustomed to 2 DOFs joystick. After the second trial, in all directions, the body contributes to more than 85\% of the task in average, showing that the initial body usage does not fade away over time, but rather increase\nj{s}.

\section{Discussion}

Body mode has demonstrated a slight improvement of the completion time compared to Joystick Mode. However, it involves significantly more physical exertion than Joystick mode, as indicated by both questionnaires responses and measured thorax displacements. Yet, in the Dual mode, where participants were free to choose between the two modes, more than 80\% of the tasks were performed using body motions. This natural inclination towards body motions might suggest an intuitive preference of the position-position body control over a third party velocity-controller, as long as the physical demands remains manageable. This observation is further supported by the fact that the body consistently initiates motion ahead of the joystick, with the joystick being used later for fine-tuning or minor adjustments. Furthermore, this behavior remains consistent over time: during the second Dual trial, the use of the body increased despite participants potentially experiencing fatigue (the experiment lasted 30 to 60 minutes depending on the participants).

These results were also observed despite participants never having trained with our interface. The training effect of the Body Mode was at least as effective as in Joystick Mode, evidenced by a significant reduction in the variance of task durations (Fig.~\ref{fig:times}), which also suggests the intuitiveness of the interface.

In line with previous research \cite{won_kim_comparison_1987, zhai_human_1995}, we observed an apparent preference for position-position control in Dual mode. Yet, in this experiment, this preference persists despite the high physical demands associated with the thorax control. However, for more extensive and energy expensive movements, the joystick comes into play to complete the motion. It is likely that the use of the joystick will become more prominent in more challenging tasks, such as executing larger effector motions or handling rotations, which can be physically demanding due to the lever effect. Hence, the utilization of an additional velocity controller would remain essential to effectively operate within a broader task space.

Additionally, it appears that this behavior is largely unconscious, as no significant differences emerged in the "intuitiveness"-related question, and control modes were rated similarly in terms of the global grade. During the Body control mode, the robotic arm was however more frequently perceived as "an extension" of the user. To draw more conclusive insights on this aspect, further studies on body representation are needed, but it may align with previous findings suggesting that tools \cite{maravita_tools_2004}, assistive devices \cite{pazzaglia_embodiment_2016} or worn robotic devices like prostheses \cite{sato_incorporation_2017} can be integrated into our body schema.

We should however note that the participants were not representative. They were healthy, relatively young (less than 45 years old) and the acceptance of such technology, as well as the mobility of the users might differ with other group of participants. Namely, the cognitive load could increase, resulting in significant results in the questionnaires. The virtual link was also constantly fixed to the participants' thorax during a trial, while a real-case scenario would repeatedly attach and detach the link for each new task, and a mechanism to attach the virtual link should be used, for example using gaze. These observations could impact the acceptance of the technology by the users, which would remain critical for a daily implementation.

Lastly, it's important to note that all of these results were observed while participants used their thorax. It is reasonable to anticipate even better performances with a hand-based interface (thus gaining in agility at a price of loosing the hand-freedom for controlling the robot). These results also suggest that such a system could be well-suited for assistive applications, namely using trunk or head motions to control the robotic arm (which can be captured through the native headset's SLAM and Inertial Measurement Units, eliminating the need for external optical motion capture). Nonetheless, while the system can facilitate a wide range of horizontal and vertical motion of the end-effector via natural body rotations, depth control requires substantial body translations, and is best handled using a velocity controller. As a result, a hybrid dual control approach appears to be the most suitable for such applications, combining position control for smaller task spaces and velocity control for broader and more physically demanding motions.

\section{Discovery sessions with impaired participants}

Two participants, each affected by upper body movement-restricting pathologies (spinal amyotrophy and partial locked-in syndrome), which restricted their head movement amplitude to less than a few centimeters of displacement, also tested the system. One of them routinely employs a Kinova Jaco arm, controlled by the manufacturer conventional 2 DoF joystick interface \cite{noauthor_jaco_nodate}, while the other participant lacked prior experience with an assistive robotic arm but uses a 2 DoF joystick to control her wheelchair. As the data gathering of such participants can contain sensitive data, and in compliance with ethical committee guidelines and French (law Jardé) and European laws governing sensitive data protection, data from these preliminary sessions were not recorded and the following feedback is qualitative. The robot itself was also far away from the participants to ensure safety.

During a half-hour session, each participant recreatively used the Hololens mode with virtual fixation to the head to perform pick-and-place tasks, specifically grasping cups or pens and placing them on a wide table. Task performance metrics were not recorded; our aim was to gather their feedback and preferences. Participants only experimented with the Body mode, excluding the Dual mode due to their limited hand motions and inability to use the joysticks available in the laboratory.

After 5 to 10 minutes of practice, using head motions, participants successfully performed various proposed tasks, especially vertically and horizontally, facilitated by the transformation of head rotation to robotic end-effector position – an achievement unattainable through direct mapping as the length of the link scales the effector's motions. It could be noted that, in contrast to position-to-velocity controllers that require tuning, the body-based control solely relies on geometric relations and, consequently, tasks were achieved without any customization. Both participants, despite distinct pathologies, were able to operate the system with little training.

\begin{figure}[t!]
\centering
  \includegraphics[width=0.8\linewidth]{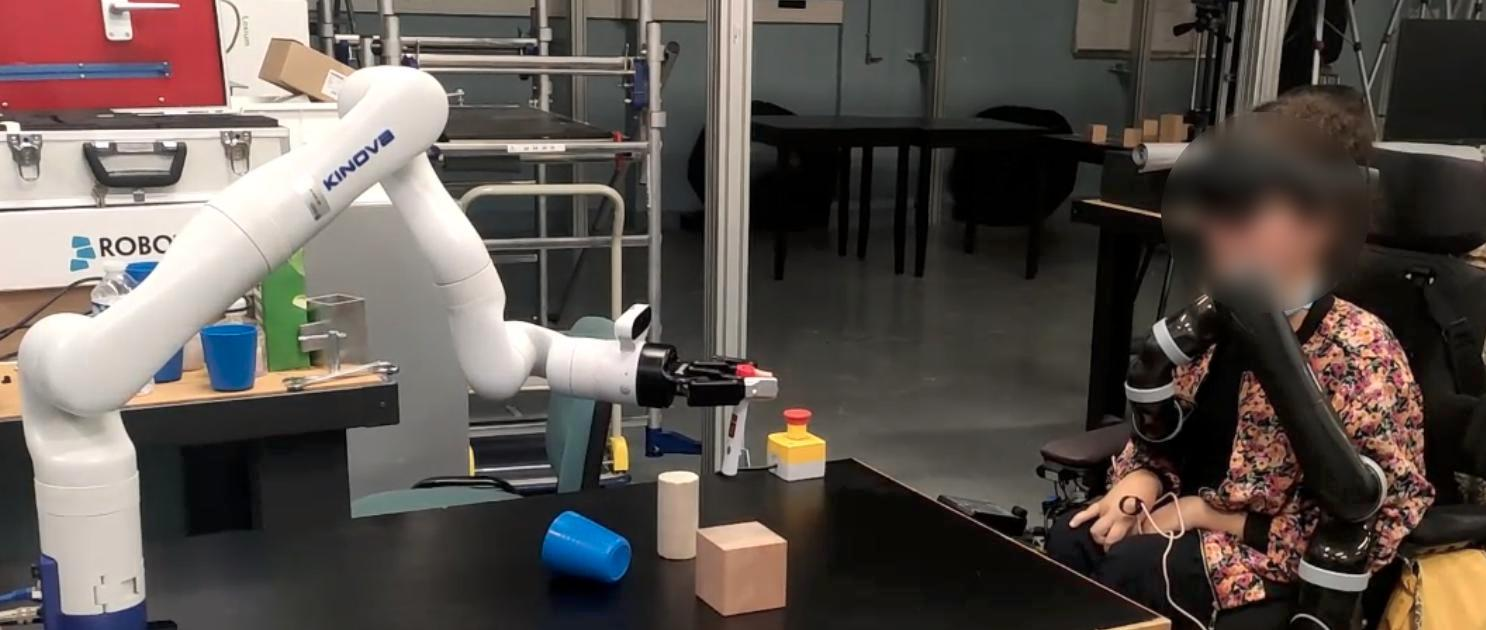}
\vspace{-2mm}
\caption{The first participant using the proposed body control. A Kinova arm is set on her wheelchair, which she use daily.}
\label{fig:assistive1}
\end{figure}
\begin{figure}[t!]
\centering
  \includegraphics[width=0.8\linewidth]{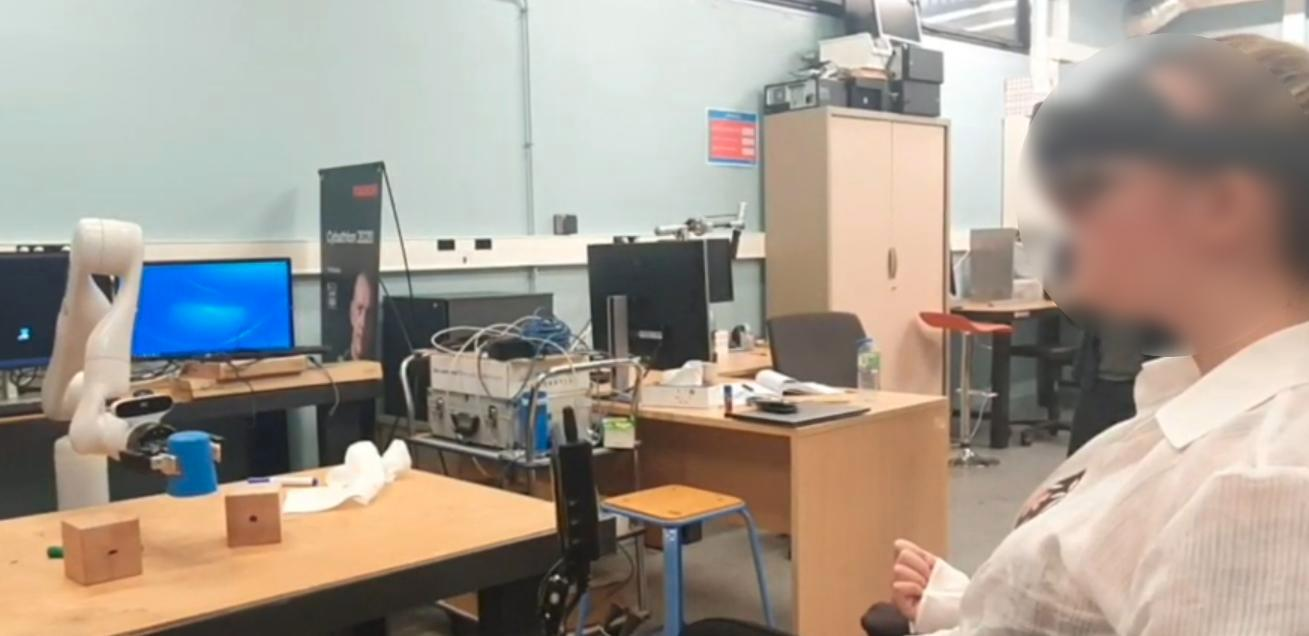}
\vspace{-2mm}
\caption{The second participant using the proposed body control. She had no prior experiences with robotics, but was familiar with gaze-to-text software.}
\label{fig:assistive2}
\end{figure}

However, depth-related tasks proved challenging due to the limited neck and torso movements of the participants. Even if they could leverage the motion of their wheelchair to control depth, such method is not viable in uncontrolled environments. The first participant suggested the potential use of their own joystick or a force sensor on their headrest for hybrid control along that direction, while the second, accustomed to gaze detectors for speech, proposed utilizing the latter. They also observed that while the headset itself was not particularly heavy, the presence of the embedded battery and electronics at the back of their head restricted the option of comfortably resting against their headrest. Consequently, the headset case should be relocated to address this constraint.

To conclude, these preliminary observations in a head-based control were promising, particularly with individuals exhibiting restricted upper body movements, and could be compared to \cite{jewell_custom-made_1989} which used rigid head-pointers to draw or type on a keyboard. Participants seemed enthusiast and curious to explore further possibilities. However, as previously mentioned, a hybrid control mode remains necessary for challenging directions, such as depth control, and would likely be essential for handling difficult orientations which where not yet explored in this paper. A hybrid coupling, integrating personalized inputs (joystick, gaze, etc.), could facilitate intentional control of more than 3 degrees of freedom simultaneously. This stands in contrast to existing sole joystick systems, which are inherently constrained by the hand dexterity of their users and usually limited to 2 DoF.

\section{Conclusion}

In this paper, we have introduced an approach for teleoperating a robotic manipulator in a nearby environment. This approach could serve to either restore lost manipulation abilities in an assistive context, or extend one's capacity to manipulate objects without relying on natural hands, potentially opening up new possibilities for tasks involving more than two hands. Our experimental results have demonstrated that this interface is intuitive to use and that there is a natural inclination to employ body motions rather than traditional external velocity joystick controllers. Furthermore, the hybrid dual control mode has been shown to combine the advantages of both position and velocity controllers without compromising performance. We plan to replicate this study with a 6 DOFs task to explore the generalization of these findings in more complex and expansive tasks. Additionally, investigating the performance of body (or dual) control with the virtual link attached to different body parts would be valuable for characterizing the effects of the location (proximal or distal) of the virtual link fixation on the body. For example, using the head might interfere with exploration of the environment or with another simultaneous task.

We emphasize that the body-to-robot link control introduced in this study is not intended to replace existing joystick solutions but rather to complement them. It offers a hand-less 3-degree-of-freedom (3 DoF) position control that requires no tuning and proves intuitive for small-range tasks, while the joystick velocity control remains applicable for larger or more challenging motions. This pointer mode stands out, to our knowledge, as the sole position control mode in the literature that is fully voluntary and potentially adaptable for use in assistive robotics.

Finally, we observed that body motions consistently initiate before joystick motions. This sequential pattern raises the possibility of partially automating joystick velocity inputs by analyzing body motions. This, at term, could eliminate the need for an external input hand-based controller to generate velocity control of the end-effector, allowing for the generation of both position and velocity control within a single postural input. Such a paradigm has been explored in the context of master-slave teleoperation \cite{dominjon_bubble_2005} and prosthesis control \cite{legrand_closing_2021} and could be applied to our approach.

\hypersetup{urlcolor = black}

\UseRawInputEncoding
\bibliography{BodyInterface.bib}

\begin{thebibliography}{10}
\providecommand{\url}[1]{#1}
\csname url@samestyle\endcsname
\providecommand{\newblock}{\relax}
\providecommand{\bibinfo}[2]{#2}
\providecommand{\BIBentrySTDinterwordspacing}{\spaceskip=0pt\relax}
\providecommand{\BIBentryALTinterwordstretchfactor}{4}
\providecommand{\BIBentryALTinterwordspacing}{\spaceskip=\fontdimen2\font plus
\BIBentryALTinterwordstretchfactor\fontdimen3\font minus
  \fontdimen4\font\relax}
\providecommand{\BIBforeignlanguage}[2]{{%
\expandafter\ifx\csname l@#1\endcsname\relax
\typeout{** WARNING: IEEEtran.bst: No hyphenation pattern has been}%
\typeout{** loaded for the language `#1'. Using the pattern for}%
\typeout{** the default language instead.}%
\else
\language=\csname l@#1\endcsname
\fi
#2}}
\providecommand{\BIBdecl}{\relax}
\BIBdecl

\bibitem{casadio_body-machine_2012}
\BIBentryALTinterwordspacing
M.~Casadio, R.~Ranganathan, and F.~A. Mussa-Ivaldi,
  ``\BIBforeignlanguage{en}{The {Body}-{Machine} {Interface}: {A} {New}
  {Perspective} on an {Old} {Theme}},'' \emph{\BIBforeignlanguage{en}{Journal
  of Motor Behavior}}, vol.~44, no.~6, pp. 419--433, Nov. 2012.
\BIBentrySTDinterwordspacing

\bibitem{jain_assistive_2015}
\BIBentryALTinterwordspacing
S.~Jain, A.~Farshchiansadegh, A.~Broad, F.~Abdollahi, F.~Mussa-Ivaldi, and
  B.~Argall, ``\BIBforeignlanguage{en}{Assistive robotic manipulation through
  shared autonomy and a {Body}-{Machine} {Interface}},'' in
  \emph{\BIBforeignlanguage{en}{2015 {IEEE} {International} {Conference} on
  {Rehabilitation} {Robotics} ({ICORR})}}.\hskip 1em plus 0.5em minus
  0.4em\relax Singapore, Singapore: IEEE, Aug. 2015, pp. 526--531.
\BIBentrySTDinterwordspacing

\bibitem{craig_wireless_2005}
\BIBentryALTinterwordspacing
D.~Craig and H.~Nguyen, ``\BIBforeignlanguage{en}{Wireless {Real}-{Time} {Head}
  {Movement} {System} {Using} a {Personal} {Digital} {Assistant} ({PDA}) for
  {Control} of a {Power} {Wheelchair}},'' in \emph{\BIBforeignlanguage{en}{2005
  {IEEE} {Engineering} in {Medicine} and {Biology} 27th {Annual}
  {Conference}}}.\hskip 1em plus 0.5em minus 0.4em\relax Shanghai, China: IEEE,
  2005, pp. 772--775.
\BIBentrySTDinterwordspacing

\bibitem{casadio_body_2011}
\BIBentryALTinterwordspacing
M.~Casadio, A.~Pressman, S.~Acosta, Z.~Danzinger, A.~Fishbach, F.~A.
  Mussa-Ivaldi, K.~Muir, H.~Tseng, and D.~Chen, ``\BIBforeignlanguage{en}{Body
  machine interface: {Remapping} motor skills after spinal cord injury},'' in
  \emph{\BIBforeignlanguage{en}{2011 {IEEE} {International} {Conference} on
  {Rehabilitation} {Robotics}}}.\hskip 1em plus 0.5em minus 0.4em\relax Zurich:
  IEEE, Jun. 2011, pp. 1--6.
\BIBentrySTDinterwordspacing

\bibitem{won_kim_comparison_1987}
\BIBentryALTinterwordspacing
{Won Kim}, F.~Tendick, S.~Ellis, and L.~Stark, ``\BIBforeignlanguage{en}{A
  comparison of position and rate control for telemanipulations with
  consideration of manipulator system dynamics},''
  \emph{\BIBforeignlanguage{en}{IEEE Journal on Robotics and Automation}},
  vol.~3, no.~5, pp. 426--436, Oct. 1987.
\BIBentrySTDinterwordspacing

\bibitem{rizzoglio_non-linear_2023}
F.~Rizzoglio, M.~Giordano, F.~A. Mussa-Ivaldi, and M.~Casadio,
  ``\BIBforeignlanguage{en}{A {Non}-{Linear} {Body} {Machine} {Interface} for
  {Controlling} {Assistive} {Robotic} {Arms}},''
  \emph{\BIBforeignlanguage{en}{IEEE TRANSACTIONS ON BIOMEDICAL ENGINEERING}},
  vol.~70, no.~7, 2023.

\bibitem{couraud_etude_2018}
M.~Couraud, ``\BIBforeignlanguage{fr}{Etude du contrôle sensorimoteur dans un
  contexte artificiel simplifié en vue d'améliorer le contrôle des
  prothèses myoélectriques.}'' 2018.

\bibitem{lincoln_visual_1953}
\BIBentryALTinterwordspacing
R.~C. Lincoln, ``\BIBforeignlanguage{en}{Visual tracking: {III}. {The}
  instrumental dimension of motion in relation to tracking accuracy.}''
  \emph{\BIBforeignlanguage{en}{Journal of Applied Psychology}}, vol.~37,
  no.~6, pp. 489--493, Dec. 1953.
\BIBentrySTDinterwordspacing

\bibitem{jagacinski_fitts_1980}
R.~J. Jagacinski, D.~W. Repperger, M.~S. Moran, S.~L. Ward, and B.~Glass,
  ``\BIBforeignlanguage{en}{Fitts' {Law} and the {Microstructure} of {Rapid}
  {Discrete} {Movements}},'' \emph{\BIBforeignlanguage{en}{Journal of
  Experimental Psychology: Human Perception and Performance}}, no. 6(2), pp.
  309--320, 1980.

\bibitem{zhai_human_1995}
S.~Zhai, ``\BIBforeignlanguage{en}{Human {Performance} in {Six} {Degree} of
  {Freedom} {Input} {Control}},'' 1995.

\bibitem{freschi_technical_2013}
\BIBentryALTinterwordspacing
C.~Freschi, V.~Ferrari, F.~Melfi, M.~Ferrari, F.~Mosca, and A.~Cuschieri,
  ``\BIBforeignlanguage{en}{Technical review of the da {Vinci} surgical
  telemanipulator: {Technical} review of the da {Vinci} surgical
  telemanipulator},'' \emph{\BIBforeignlanguage{en}{The International Journal
  of Medical Robotics and Computer Assisted Surgery}}, vol.~9, no.~4, pp.
  396--406, Dec. 2013.
\BIBentrySTDinterwordspacing

\bibitem{pulgarin_assessing_nodate}
E.~J.~L. Pulgarin, O.~Tokatli, G.~Burroughes, and G.~Herrmann,
  ``\BIBforeignlanguage{en}{Assessing tele-manipulation systems using task
  performance for glovebox operations},''
  \emph{\BIBforeignlanguage{en}{Frontiers in Robotics and AI}}, 2022.

\bibitem{dominjon_comparison_nodate}
L.~Dominjon, A.~Lécuyer, J.-M. Burkhardt, and R.~Simon,
  ``\BIBforeignlanguage{en}{A {Comparison} of {Three} {Techniques} to
  {Interact} in {Large} {Virtual} {Environments} {Using} {Haptic} {Devices}
  with {Limited} {Workspace}}.''

\bibitem{noauthor_headpointer_2017}
\BIBentryALTinterwordspacing
``\BIBforeignlanguage{en-US}{Headpointer - {Handi} {Life} {Sport}},'' Oct.
  2017. [Online]. Available:
  \url{https://handilifesport.com/product/headpointer/}
\BIBentrySTDinterwordspacing

\bibitem{jewell_custom-made_1989}
K.~H. Jewell, ``A custom-made head pointer for children,'' \emph{The American
  Journal of Occupational Therapy}, vol.~43, no.~7, pp. 456--460, 1989,
  publisher: The American Occupational Therapy Association, Inc.

\bibitem{goodman_computer_2008}
\BIBentryALTinterwordspacing
N.~Goodman, A.~M. Jette, B.~Houlihan, and S.~Williams,
  ``\BIBforeignlanguage{en}{Computer and {Internet} {Use} by {Persons} {After}
  {Traumatic} {Spinal} {Cord} {Injury}},''
  \emph{\BIBforeignlanguage{en}{Archives of Physical Medicine and
  Rehabilitation}}, vol.~89, no.~8, pp. 1492--1498, Aug. 2008.
\BIBentrySTDinterwordspacing

\bibitem{khoramshahi_practical_2023}
M.~Khoramshahi, A.~Poignant, G.~Morel, and N.~Jarrassé, ``A {Practical}
  {Control} {Approach} for {Safe} {Collaborative} {Supernumerary} {Robotic}
  {Arms},'' 2023.

\bibitem{noauthor_spacemouse_nodate}
\BIBentryALTinterwordspacing
``\BIBforeignlanguage{en-US}{{SpaceMouse} {Compact} - {Official} {3Dconnexion}
  {UK} store}.'' [Online]. Available:
  \url{https://3dconnexion.com/us/product/spacemouse-compact/}
\BIBentrySTDinterwordspacing

\bibitem{gonzalez_measurement_2010}
\BIBentryALTinterwordspacing
Ã.~González, ``\BIBforeignlanguage{en}{Measurement of areas on a sphere using
  {Fibonacci} and latitude-longitude lattices},''
  \emph{\BIBforeignlanguage{en}{Mathematical Geosciences}}, vol.~42, no.~1, pp.
  49--64, Jan. 2010, arXiv:0912.4540 [math].
\BIBentrySTDinterwordspacing

\bibitem{maravita_tools_2004}
\BIBentryALTinterwordspacing
A.~Maravita and A.~Iriki, ``\BIBforeignlanguage{en}{Tools for the body
  (schema)},'' \emph{\BIBforeignlanguage{en}{Trends in Cognitive Sciences}},
  vol.~8, no.~2, pp. 79--86, Feb. 2004.
\BIBentrySTDinterwordspacing

\bibitem{pazzaglia_embodiment_2016}
\BIBentryALTinterwordspacing
M.~Pazzaglia and M.~Molinari, ``\BIBforeignlanguage{en}{The embodiment of
  assistive devices—from wheelchair to exoskeleton},''
  \emph{\BIBforeignlanguage{en}{Physics of Life Reviews}}, vol.~16, pp.
  163--175, Mar. 2016.
\BIBentrySTDinterwordspacing

\bibitem{sato_incorporation_2017}
\BIBentryALTinterwordspacing
Y.~Sato, T.~Kawase, K.~Takano, C.~Spence, and K.~Kansaku,
  ``\BIBforeignlanguage{en}{Incorporation of prosthetic limbs into the body
  representation of amputees: {Evidence} from the crossed hands temporal order
  illusion},'' in \emph{\BIBforeignlanguage{en}{Progress in {Brain}
  {Research}}}.\hskip 1em plus 0.5em minus 0.4em\relax Elsevier, 2017, vol.
  236, pp. 225--241.
\BIBentrySTDinterwordspacing

\bibitem{noauthor_jaco_nodate}
\BIBentryALTinterwordspacing
``\BIBforeignlanguage{en}{Jaco® assistive robot {User} guide}.''
\BIBentrySTDinterwordspacing

\bibitem{dominjon_bubble_2005}
\BIBentryALTinterwordspacing
L.~Dominjon, A.~Lecuyer, J.~Burkhardt, G.~Andrade-Barroso, and S.~Richir,
  ``\BIBforeignlanguage{en}{The "{Bubble}" {Technique}: {Interacting} with
  {Large} {Virtual} {Environments} {Using} {Haptic} {Devices} with {Limited}
  {Workspace}},'' in \emph{\BIBforeignlanguage{en}{First {Joint} {Eurohaptics}
  {Conference} and {Symposium} on {Haptic} {Interfaces} for {Virtual}
  {Environment} and {Teleoperator} {Systems}}}.\hskip 1em plus 0.5em minus
  0.4em\relax Pisa, Italy: IEEE, 2005, pp. 639--640.
\BIBentrySTDinterwordspacing

\bibitem{legrand_closing_2021}
\BIBentryALTinterwordspacing
M.~Legrand, N.~Jarrasse, E.~d. Montalivet, F.~Richer, and G.~Morel,
  ``\BIBforeignlanguage{en}{Closing the {Loop} {Between} {Body} {Compensations}
  and {Upper} {Limb} {Prosthetic} {Movements}: {A} {Feasibility} {Study}},''
  \emph{\BIBforeignlanguage{en}{IEEE Transactions on Medical Robotics and
  Bionics}}, vol.~3, no.~1, pp. 230--240, Feb. 2021.
\BIBentrySTDinterwordspacing

\end{thebibliography}
\bibliographystyle{IEEEtran}

\end{document}